\documentclass{article} 
\usepackage{iclr2024_conference,times}

\usepackage{amsmath,amsfonts,bm}

\def\eqref#1{equation~\ref{#1}}

\DeclareMathAlphabet{\mathsfit}{\encodingdefault}{\sfdefault}{m}{sl}
\SetMathAlphabet{\mathsfit}{bold}{\encodingdefault}{\sfdefault}{bx}{n}

\usepackage{hyperref}
\usepackage{url}

\title{{ALMA: Alignment with Minimal Annotation}}

\author{\qquad Michihiro Yasunaga, ~ Leonid Shamis, ~ Chunting Zhou, ~ Andrew Cohen,\\[0.5ex]
\qquad\qquad\textbf{Jason Weston, ~ Luke Zettlemoyer, ~ Marjan Ghazvininejad}
\\[1.5ex]
\hspace{17em}Meta FAIR
}

\usepackage{inconsolata}
\usepackage{graphicx}
\usepackage{xcolor}
\usepackage{color, colortbl}
\usepackage{multirow}
\usepackage{makecell}
\usepackage{arydshln}
\usepackage{booktabs}
\usepackage{listings}
\usepackage{xspace}
\usepackage{wrapfig}
\usepackage{changes}
\usepackage[nomessages]{fp}
\newcommand{\methodname}{ALMA\xspace}

\makeatletter
\renewcommand\paragraph{\@startsection{paragraph}{4}{\z@}{0.001ex plus 0.001ex minus .001ex}{-1em}{\normalsize\bf}}
\makeatother

\definecolor{RoyalBlue}{HTML}{0272bb}
\hypersetup{
  colorlinks   = true, 
  urlcolor     = RoyalBlue, 
  linkcolor    = RoyalBlue, 
  citecolor   = RoyalBlue 
}

\iclrfinalcopy 
\begin{document}

\maketitle

\begin{abstract}
Recent approaches to large language model (LLM) alignment typically require millions of human annotations or rely on external aligned models for synthetic data generation. This paper introduces \textbf{ALMA: Alignment with Minimal Annotation}, demonstrating that effective alignment can be achieved using only 9,000 labeled examples---less than 1\% of conventional approaches. ALMA generates large amounts of high-quality synthetic alignment data through new techniques: diverse prompt synthesis via few-shot learning, diverse response generation with multiple model checkpoints, and judge (reward model) enhancement through score aggregation and self-distillation. Using only a pretrained Llama3 base model, 5,000 SFT examples, and 4,000 judge annotations, ALMA achieves performance close to Llama3-Instruct across diverse alignment benchmarks (e.g., 0.1\% difference on AlpacaEval 2.0 score). These results are achieved with a multi-round, self-bootstrapped data synthesis and training recipe that continues to improve for 10 rounds, surpassing the typical 3-round ceiling of previous methods. These results suggest that base models already possess sufficient knowledge for effective alignment, and that synthetic data generation methods can expose it.
\end{abstract}

\begin{figure}[h!]
    \centering
    \includegraphics[width=1.01\linewidth]{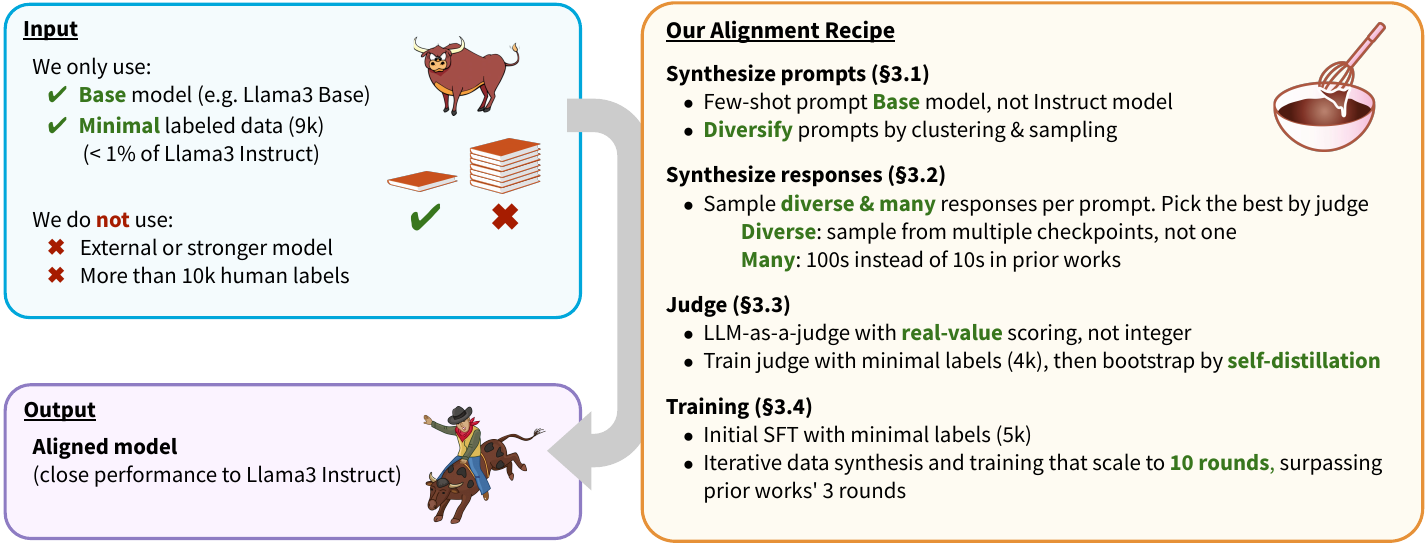}
    \caption{\textbf{Overview of our approach \methodname: Alignment with Minimal Annotation}.
    Starting with only a pretrained base LLM (Llama3 Base) and minimal seed data (9k samples---less than 1\% of conventional approaches), we align the model to achieve performance close to Llama3 Instruct (left panel). This is achieved through our new alignment techniques (right panel) that enhance each of the four key components in alignment: prompt synthesis (\S \ref{sec:method-prompt}), response synthesis (\S \ref{sec:method-response}), judge (\S \ref{sec:method-judge}), and model training (\S \ref{sec:method-train}). 
    }  
    \label{fig:overview} 
\end{figure}

\section{Introduction}

Language model alignment is an important step to train models to respond to user instructions accurately and helpfully, but can require a significant amount of data curation. It typically consists of supervised finetuning (SFT) and learning from feedback, which involve solving four sub-problems:
(1) Prompt collection: gathering user instructions and queries;
(2) Response collection: first collecting human-annotated responses to these prompts (SFT data), then generating model responses;
(3) Judge: developing judges or reward models to assess response quality, typically trained on human preference data; and
(4) Model training: aligning the LLM using the collected prompts, responses, and judgments through optimization algorithms such as SFT, PPO \citep{instructgpt}, and DPO \citep{rafailov2023dpo}.
Extensive human-annotated data is often used for all components---prompts, responses, and judgments. For example, state-of-the-art aligned LLMs such as Llama3 Instruct \citep{llama3} have used millions of human annotations during their alignment training.

In response to this challenge, recent research has shown that it is possible to align models on synthetically generated data \citep{xu2024magpie, alpaca, chen2023alpagasus, gulcehre2023reinforced}.
However, these approaches typically rely on external models for data synthesis, which themselves were aligned on large amounts of (typically undisclosed) labeled data. 

In this work, we argue that the \textbf{base model itself} already has sufficient knowledge to synthesize high-quality training data, requiring only minimal seed annotations for bootstrapping. 
We show that the model's alignment capabilities can be surfaced and reinforced through careful data synthesis techniques and continual training.

To test this hypothesis, we use orders of magnitude less human-labeled data (9k instead of the conventional millions) and use the model itself as the only source of synthetic data generation.
As shown in Figure \ref{fig:overview}, our approach performs model alignment using only three inputs: a pretrained base LLM (Llama3 base), a small seed SFT dataset (5k), and a small judge dataset (4k)---without requiring distillation from additional models or datasets. Our approach, termed \textbf{ALMA (Alignment with Minimal Annotation)}, comprises the following techniques:
\begin{itemize}
    \setlength{\leftskip}{-30pt}
    \item[] (1) Prompt synthesis: We synthesize new prompts through few-shot prompting of the base model, then apply clustering and subsampling to enhance prompt diversity.
    \item[] (2) Response synthesis: To collect high-quality responses, we generate diverse outputs using multiple model checkpoints (rather than a single checkpoint) and produce 100s of responses per prompt (instead of the 10s typically used in prior works). The judge then selects the best responses.
    \item[] (3) Judge: We initially finetune the LLM as a judge using only a small labeled dataset (4k). To enhance performance, we enable the LLM judge to produce real-value scores instead of the conventional integer scores, and further train it via self-distillation using additional synthetic prompts and responses. 
    \item[] (4) Model training: Following initial SFT with a small seed dataset (5k), we train the LLM using the synthesized data over 10 rounds.
\end{itemize}
ALMA relies solely on the base model and small seed datasets, with no distillation from external models. This self-bootstrapping approach is made possible by our data synthesis techniques that improve the quality and diversity of prompts and responses as well as the performance of the judge.

This overall approach leads to qualitatively different learning trends, enabling more iterations of training on higher quality synthetic data. 
ALMA allows {\em sustained improvement} where the model continually improves for 10 rounds, while previous alignment methods typically showed 3 rounds of improvements (\S \ref{sec:main-result}).
We also show that sampling many (100s) responses using multiple model checkpoints significantly improves performance compared to previous methods, which typically sample up to 10--20 responses using a single model (\S \ref{sec:analysis-response}). 

In our evaluation on diverse alignment benchmarks, including MT-Bench, Arena-Hard, and AlpacaEval, our final model achieved close performance to Llama3-Instruct (e.g., overall average score of 8.22 vs 8.25 on the GPT-4o evaluator \citep{zheng2023judging}), while using only 9k labeled examples---less than 1\% of Llama3-Instruct (Figure \ref{fig:result_curve}). 

These results show that base language models possess sufficient latent knowledge to bootstrap their own alignment process from minimal seed data, dramatically reducing the need for extensive human annotation. While high-quality seed data remains essential---suggesting that human input cannot be eliminated entirely---ALMA presents a fundamental shift from conventional alignment approaches that depend on millions of labeled examples. This work opens important research directions, including identifying what types of human annotations are most valuable for bootstrapping and inherently difficult to synthesize, and understanding the complementary roles of synthetic and human-annotated data. ALMA suggests that future alignment methods may benefit not only from ever-larger human-annotated datasets but also from sophisticated data synthesis techniques.

\section{Related works}
Since the early success of RLHF-based LLM alignment \citep{instructgpt, bai2022constitutional}, extensive research has been done to improve open-source recipes for LLM alignment, including Llama3 \citep{llama3}, RLHF Workflow \citep{dong2024rlhf}, and Nemotron \citep{adler2024nemotron}. While these works achieve impressive results, they typically rely on large amounts of human-annotated data, ranging from tens of thousands (e.g., Nemotron) to millions (e.g., Llama3). In contrast, our current work aims to use minimum human-annotated data (below 10k).

Given the high cost of human annotation, recent research has explored synthetic data generation as a way to reduce reliance on human-labeled data \citep{xu2024magpie, yuan2024selfrewarding, adler2024nemotron,li2024selfalignment,tong2024dartmath, gulcehre2023reinforced, chen2023alpagasus}. For instance, Magpie \citep{xu2024magpie} shows that high-quality SFT data can be synthesized by simply prompting Llama3-Instruct. 
Nemotron \citep{adler2024nemotron} presents methods to synthesize RLHF data, including prompts, responses and preference labels.
However, these studies typically assume access to external models (LLMs that have been aligned or trained with more data, such as Llama3-Instruct and Mixtral-Instruct) for data synthesis, which may indirectly depend on additional human annotation. In contrast, our approach uses only the \textbf{base} model and a small seed dataset for data synthesis, and does not rely on additional annotation.

Our work is most closely related to Self-Rewarding LM \citep{yuan2024selfrewarding}, which primarily uses a base model for self-alignment. We build upon their approach and develop an improved recipe that enables a base model to fully self-bootstrap through several technical enhancements: high-quality data synthesis through diverse prompt and response sampling (\S \ref{sec:method-prompt}, \ref{sec:method-response}); self-distillation for further improving the judge (\S \ref{sec:method-judge}); and an iterative training recipe that allows the model to improve beyond 5 rounds without overfitting (\S \ref{sec:method-train}).

Another line of related works is label-efficient SFT for alignment \citep{sun2024principle, wang2022selfinstruct, zhou2024lima}. For example, LIMA \citep{zhou2024lima} demonstrated the viability of minimal training data for SFT. Our work extends this line of research to cover the full alignment pipeline, including both SFT and learning from feedback. 

Finally, our work is related to works that invest extra compute in LLM inference \citep{wei2022chain, kojima2022large, openai_o1}. While the typical goal is to produce better responses for users, our work uses these improved responses for model training (i.e., feeding them back to the model to reinforce positive behaviors). By doing so, we do not need to use additional inference resources when serving the model to users.

\section{Approach}
\label{sec:method}
\paragraph{Preliminary.} The alignment process typically consists of four key components \citep{yuan2024selfrewarding}:
\begin{itemize}
\setlength{\leftskip}{-30pt}
\item[] (1) Prompt collection: gathering instructions or queries that users ask LLMs
\item[] (2) Response collection: obtaining human-annotated responses (SFT data) or model-generated responses to these prompts
\item[] (3) Judge: creating judges or reward models to evaluate response quality given a prompt, typically through training on human-annotated preference data (judge data)
\item[] (4) Model training: aligning the LLM through a two-step process---(i) first using human-annotated responses (SFT), and (ii) then using model-generated responses and judgments via optimization algorithms such as best-of-n SFT, PPO or DPO (feedback learning). The feedback learning phase can be repeated for multiple rounds by generating new responses with the latest model and incorporating this fresh data into subsequent training rounds.
\end{itemize}
Traditional alignment approaches use millions of human-annotated samples for SFT and judge data (e.g., \citealt{llama3}) or rely on external models to synthesize such data (e.g., \citealt{adler2024nemotron}).

\textbf{In this work}, we present ALMA, a method to achieve alignment performance close to Llama3 Instruct by using only three minimal components---\textbf{a base model}: Llama3 Base, \textbf{minimal SFT data}: e.g., 5k prompt-response samples from an open-source SFT dataset \citep{wang2024helpsteer2}, and \textbf{minimal judge data}: e.g., 4k judgment samples we annotate in \S \ref{sec:method-judge-data}. 
We achieve this result through self-bootstrapping data synthesis and training, with enhanced techniques across the four key components: improved data synthesis for prompts (\S \ref{sec:method-prompt}) and responses (\S \ref{sec:method-response}), along with refined training methods for the judge (\S \ref{sec:method-judge}) and the model (\S \ref{sec:method-train}). We elaborate on each of these components below.

\subsection{Prompt Synthesis}
\label{sec:method-prompt}
Collecting high-quality prompts---those that are diverse and representative of real user queries to LLMs---is important for alignment training. While previous works relied on human annotation to collect millions of prompts, we explore using LLMs to synthesize prompts.

Our basic synthesis method builds upon the self-instruct technique \citep{wang2022selfinstruct}. Specifically, we randomly sample 3-5 prompts from our 5k seed prompts as context, and then few-shot prompt the base LLM to generate a new prompt. After repeating this process and removing duplicates, we obtained a set of 100M unique synthetic prompts in total, which can be sampled for each round of feedback learning (\S \ref{sec:method-train}).

While this approach generates realistic prompts through few-shot conditioning, it has limitations. The generated prompts skew toward common queries (e.g., "Plan a three-day trip to Seattle") and underrepresent longer, complex, or domain-specific queries (e.g., specific medical questions with detailed symptoms and lab results).
Simply sampling prompts uniformly at random from this set for training would overemphasize common prompts and potentially harm performance on more challenging ones. To address this, we developed a \textbf{diversified sampling} technique:

Our large set of 100M prompts does contain long-tail prompts, albeit in smaller numbers. To better utilize these diverse examples, we implement clustering-based rebalancing. We perform k-means clustering on the 100M prompts, creating 1M clusters based on the hidden representations from the base LLM. During each round of feedback learning---which uses $K$ prompts (e.g., $K$=50,000)---we select $K$ clusters and sample one previously unused prompt from each cluster. This approach ensures a diverse selection of prompts from the original set. We show the effectiveness of this diversification technique in Table \ref{tab:ablation_prompt}.

\subsection{Response Synthesis}
\label{sec:method-response}
After the initial SFT phase that uses human-labeled responses, feedback learning is performed using model-generated responses to further enhance model performance.

Several optimization methods exist for feedback learning, including:
\begin{itemize}
\setlength{\leftskip}{-26pt}
\setlength{\itemsep}{-0pt}
\item Best-of-n SFT: sample $N$ model-generated responses per prompt, select the best ones using the judge, and then use the selected responses for additional SFT
\item PPO \citep{instructgpt}: reinforcement learning using reward scores from the judge
\item DPO \citep{rafailov2023dpo} and SimPO \citep{meng2024simpo}, etc: learning from comparative preferences between `chosen' and `rejected' responses as determined by the judge
\end{itemize}
We primarily use the best-of-n SFT method in this work due to its simplicity and effectiveness, but we will show in \S \ref{sec:analysis-response} that our approach also works well with DPO-style methods.

The quality of synthesized responses is important across all these training methods, whether selecting the best response in best-of-n or the `chosen' response in DPO. Conventional approaches typically use a single model (the latest checkpoint in training) to generate up to 10-20 responses \citep{yuan2024selfrewarding, dong2024rlhf}, but we find that significantly better responses can be obtained in best-of-n sampling by increasing both the quantity and diversity of responses---an improvement measured by both our judge's scores and extrinsic reward scores from the GPT-4o judge \citep{zheng2023judging} and Armo \citep{ArmoRM}.

Specifically, we improved response quality through two key changes:
\begin{itemize}
\setlength{\leftskip}{-26pt}
\setlength{\itemsep}{-0pt}
\item \textbf{Increased sampling}: Sample 200 responses instead of tens (effectiveness shown in Figure \ref{fig:best_of_n}).
\item \textbf{Enhanced diversity}: Sample from multiple model checkpoints (e.g., two initial SFT checkpoints trained with different seeds, and the latest checkpoint) rather than a single checkpoint. This prevents overfitting (see Table \ref{tab:ablation_multi}). We found a sampling ratio of 1:1:1 from these three checkpoints worked well.
\end{itemize}
Note that for fair comparison, we maintained a fixed total of 200 sampled responses per prompt, regardless of whether sampling from single or multiple checkpoints. 
Our final recipe incorporates this scaled and diverse response synthesis method.

\subsection{Judge}
\label{sec:method-judge}
A judge takes a prompt and a response as input to produce a score indicating response quality. Common approaches to creating judges include:
(1) Reward modeling \citep{stiennon2020learning, instructgpt}, which adds a regression head to a base LLM and trains it to produce real-valued scores, and
(2) LLM-as-a-judge \citep{zheng2023judging}, which leverages the LLM's text generation ability by prompting it to produce a judgment score, typically in the form of an integer string.
We adopt the LLM-as-a-judge approach as it requires less training data compared to training a reward model.

A baseline approach of LLM-as-a-judge would be to few-shot prompt the base LLM to produce an integer score from 0 to 10 \citep{zheng2023judging}. However, this approach yields suboptimal performance. To improve the judge performance, we implement three techniques: \textbf{high-quality seed data} for finetuning (\S \ref{sec:method-judge-data}), \textbf{fine-grained real-value score prediction} instead of integer scores (\S \ref{sec:method-judge-score}), and \textbf{self-distillation} to further bootstrap the judge training (\S \ref{sec:method-judge-distill}).

\subsubsection{Seed data for finetuning}
\label{sec:method-judge-data}
While base LLMs possess extensive knowledge and can be prompted to act as judges, fine-tuning them with even modest amounts of judgment task data (a few thousand examples) significantly improves their performance.

Existing finetuning data for LLM-as-a-judge, such as \citet{yuan2024selfrewarding}, typically used a 1--5 rating scale, which we found insufficient for differentiating between multiple responses. We therefore conducted our own human annotation using a 0-10 scale. Using 2k seed prompts, we generated two responses per prompt using Llama models and had human annotators rate each response based on the 0--10 scoring criteria, with each score labeled with an English definition, such as 0: unusable and 10: outstanding.
For complete annotation guidelines, see Appendix \ref{sec:app-judge-annotation}. 

This process yielded 4k prompt-response-score triplets, which we used to finetune the base LLM. The concrete prompt template used for LLM-as-a-judge is provided in Appendix \ref{sec:app-judge-prompt}.

\subsubsection{LLM-as-a-Judge with real-value scoring}
\label{sec:method-judge-score}
Existing LLM-as-a-judge approaches prompt LLMs to generate an integer score, e.g., 1--5 or 0--10 \citep{zheng2023judging, yuan2024selfrewarding}. However, this limited range of possible scores makes it challenging to differentiate between responses when evaluating many responses (e.g., even with the 0--10 scale, multiple responses might receive a score of 9, making it difficult to determine the best one).
To address this limitation, we look at the model's predicted probability for each possible score 0,1,...,10, and then aggregate them by taking the probability-weighted average to produce a real-valued score (e.g., 9.21). This technique proves extremely effective, improving judge performance by 17 points on RewardBench (see Table \ref{tab:ablation_judge}).

An alternative, effective way to obtain a real-value score is to prompt and sample a score from the LLM multiple times and averaging them, but the token probability aggregation approach is more computationally efficient.

\subsubsection{Self-distillation}
\label{sec:method-judge-distill}
We extend the judge's training beyond the initial 4k seed samples through synthetic data generation. Specifically, given our judge developed through \S \ref{sec:method-judge-data}--\ref{sec:method-judge-score}, we apply it to 20k synthetic prompt-response pairs generated in \S \ref{sec:method-prompt}--\ref{sec:method-response} to obtain judgment scores. These newly generated 20k judgment examples are then used to continually train our judge. As this process only uses the judge itself for data synthesis, we call it self-distillation. We find that this technique can further boost the judge's performance (Table \ref{tab:ablation_judge}).

\subsubsection{Vanilla prompting vs chain-of-thought prompting}
Another design choice for LLM-as-a-judge is prompting strategy, e.g., whether to produce the score directly (vanilla prompting) or to produce intermediate reasoning steps before producing the score (chain-of-thought prompting \citep{cot_wei, kojima2022large}). In this work, we chose vanilla prompting (concrete prompt template in Appendix \ref{sec:app-judge-prompt}) for two main reasons:
\begin{itemize}
    \setlength{\leftskip}{-26pt}
    \setlength{\itemsep}{0pt}
    \item In our finetuning experiment (\S \ref{sec:method-judge-data}), vanilla prompting performed comparably with chain-of-thought prompting 
    \item Real-value scoring (\S \ref{sec:method-judge-score}) proved extremely effective, and it integrates seamlessly with vanilla prompting. Implementing real-value scoring with chain-of-thought prompting would require sampling multiple reasoning chains and aggregating them, which is computationally expensive.
\end{itemize}
We leave further exploration of chain-of-thought prompting (e.g., how to efficiently combine it with real-value scoring) to future work.

\subsection{Model training}
\label{sec:method-train}
Using the 100M collected prompts $P$ from \S \ref{sec:method-prompt}, the response synthesis technique from \S \ref{sec:method-response}, and the developed judge $J$ from \S \ref{sec:method-judge}, we perform iterative alignment training for the model over $R$ rounds through the following algorithm:
\begin{itemize}
    \setlength{\leftskip}{-26pt}
    \item Define $M_0$ as the base model (Llama3 8B Base)
    \item Define $D_1$ as the 5k seed SFT data
    \item \textbf{Initial SFT}: Finetune the base model on $D_1$ to obtain initial SFT models $M_1^{(i)}$. We use two random seeds in training to obtain two checkpoints $M_1^{(1)}$ and $M_1^{(2)}$, which will later be used in diverse response sampling.
    \item \textbf{Learning from feedback}: For each round $r$ ($r=2,3,...,R$):
    \begin{itemize}
        \setlength{\leftskip}{-40pt}
        \item[] \textit{Data synthesis}:
        \begin{itemize}
            \setlength{\leftskip}{-44pt}
            \item \!Prompts: sample $K$ (=50k) new prompts from $P$ using the diverse sampling technique (\S \ref{sec:method-prompt}).
            \item \!Responses: for each prompt, sample $N$ (=200) responses in total using the initial SFT checkpoints $M_1^{(i)}$ and the latest checkpoint $M_{r-1}$ if $r \!\ge\! 3$
            (diverse sampling technique from \S \ref{sec:method-response}).
            \item \!Judgments: for each prompt, apply the judge $J$ to all generated responses and select the best one (best-of-n). 
            \item \!Define $D_{r}$ as the resulting set of $K$ prompt-response pairs
            \vspace{4pt}
        \end{itemize}
        \item[] \textit{Training}:
        \begin{itemize}
            \setlength{\leftskip}{-44pt}
            \item \!Finetune the base model $M_0$ on the combined datasets $D_1, ..., D_r$ to obtain the latest model $M_r$ (best-of-n SFT)
        \end{itemize}
    \end{itemize}
\end{itemize}

A key challenge in alignment training is that model performance tends to plateau after a few rounds. Existing works typically show improvements for 3 rounds \citep{dong2024rlhf}.
To prevent such plateauing, we identified several effective techniques (incorporated in the algorithm above) that helped our model improve over 10 rounds and achieve performance close to Llama3 Instruct:
\begin{itemize}
    \setlength{\leftskip}{-26pt}
    \setlength{\itemsep}{-0pt}
    \item \textbf{Sample responses from multiple checkpoints}, not just the latest one, as discussed in \S \ref{sec:method-response} and \S \ref{sec:analysis-response}. Using only the latest checkpoint may lead to overfitting and reinforce hallucinated outputs.
    \item In each round of data synthesis ($D_r$), \textbf{introduce previously unused prompts}. When generating responses with the latest model, use these new prompts rather than reusing old ones.
    This prevents overfitting, because generating responses for previously seen prompts would lack diversity as the model has already learned selected responses to them during earlier training rounds.
    \item In each round of best-of-n SFT, \textbf{initialize training from the base model checkpoint} rather than the latest checkpoint, as the latest checkpoint may have overfit and exhibit forgetting compared to the base model. Since we include all previous rounds of data $D_1, ..., D_{r-1}$ in each round ($r$) of best-of-n SFT, training from the base model still incorporates all of our synthesized data.
\end{itemize}

In these feedback learning rounds, we focus on the best-of-n SFT approach for its simplicity and effectiveness, but we will show in \S \ref{sec:analysis-response} that DPO-style approaches can be used effectively too.

\section{Experiments}

\subsection{Experimental setup}
\paragraph{Seed data.}
For the seed SFT data (prompt-response pairs), we use 5k samples from the Daring-Anteater dataset \citep{wang2024helpsteer2}. 
For the seed judge training data, we use the 4k samples collected in our human annotation process (\S \ref{sec:method-judge}), which consists of 2k prompts and judgment scores for 2 responses per prompt. Thus, our approach only uses 9k seed labeled data in total.

\paragraph{Model.}
We use Llama3 8B Base (the pretrained model) as the base model. Starting from this model, we study how closely we can reach the performance of Llama3 8B Instruct (the officially released model aligned with millions of human annotations) while using only minimal seed data.

\paragraph{Hyperparameters.}
In our alignment recipe (\S \ref{sec:method}), we let the response sampling size per prompt ($N$) be 200, the number of prompts per training round ($K$) be 50k, and the number of training rounds ($R$) be 10. These numbers were chosen based on our ablation studies detailed in \S \ref{sec:analysis}.

\paragraph{Evaluation.}
To evaluate the aligned models, we follow the evaluation protocol of existing literature on LLM alignment \citep{yuan2024selfrewarding, adler2024nemotron}.
We use diverse alignment benchmarks, including LIMA \citep{zhou2024lima} (300 prompts), MT-Bench \citep{zheng2023judging} (80 prompts), Self-Reward (SR) \citep{yuan2024selfrewarding} (250 prompts), Arena-Hard \citep{li2024crowdsourced} (500 prompts), and AlpacaEval \citep{alpaca} (600 prompts). These benchmarks consist of user prompts covering a wide range of topics and domains, including writing, role play, humanities, coding, reasoning, and STEM. We evaluate the model generated responses given these sets of prompts. Specifically, for the evaluation metrics, we use GPT-4o \citep{openai2023gpt4} and the Armo reward model \citep{ArmoRM}, which is a top-performing open-source evaluator on the RewardBench leaderboard \citep{lambert2024rewardbench}, to score the responses. As running Armo is much cheaper than GPT-4o, we use Armo score as the primary metric for the development process and ablation studies, while also reporting the GPT-4o score in the final evaluation.

Additionally, to conduct intrinsic evaluation of the judge we develop (\S \ref{sec:method-judge}), we use the RewardBench \citep{lambert2024rewardbench}, a standard method for benchmarking judges or reward models.

\paragraph{Compute.} 
We used 512 H100 GPUs for both data synthesis (prompt, response, and judgment synthesis) and model training (initial SFT and best-of-n SFT) process. The whole pipeline of data synthesis and training over $R=10$ rounds took approximately 5 days.

\subsection{Main results}
\label{sec:main-result}

\begin{table}[h]
    \newcolumntype{G}{>{\columncolor[gray]{0.9}}c} 

    \hspace{-8pt}
    \scalebox{0.79}{
    \begin{tabular}{lGccccccGccccc}
    \toprule
    \multirow{2}{*}{\begin{tabular}{@{}l@{}}\vrule width 0pt depth 0pt height 8pt 
    \textbf{Model}\end{tabular}} & \multicolumn{6}{c}{\textbf{Metric: GPT-4o score}} & \!\!\! & \multicolumn{6}{c}{\textbf{Metric: Armo score}} \\
    \cmidrule{2-7}  \cmidrule{9-14}
    & \begin{tabular}{@{}c@{}}\vrule width 0pt depth 0pt height 8pt 
    {\small Avg.}\end{tabular}
    & \!\begin{tabular}{@{}c@{}}\vrule width 0pt depth 0pt height 8pt 
    {\small LIMA}\end{tabular}\!
    & \!\begin{tabular}{@{}c@{}}\vrule width 0pt depth 0pt height 8pt 
    {\small MT}\\[-1pt]{\small Bench}\end{tabular}\!
    & \!\!\begin{tabular}{@{}c@{}}\vrule width 0pt depth 0pt height 8pt 
    {\small SR}\end{tabular}\!\!
    & \!\begin{tabular}{@{}c@{}}\vrule width 0pt depth 0pt height 8pt 
    {\small Arena}\\[-1pt]{\small Hard}\end{tabular}\!
    & \!\!\!\!\begin{tabular}{@{}c@{}}\vrule width 0pt depth 0pt height 8pt 
    {\small Alpaca}\end{tabular}\!\!\!\!
    &\!\!\!
    & \begin{tabular}{@{}c@{}}\vrule width 0pt depth 0pt height 8pt 
    {\small Avg.}\end{tabular}
    & \!\begin{tabular}{@{}c@{}}\vrule width 0pt depth 0pt height 8pt 
    {\small LIMA}\end{tabular}\!
    & \!\begin{tabular}{@{}c@{}}\vrule width 0pt depth 0pt height 8pt 
    {\small MT}\\[-1pt]{\small Bench}\end{tabular}\!
    & \!\!\begin{tabular}{@{}c@{}}\vrule width 0pt depth 0pt height 8pt 
    {\small SR}\end{tabular}\!\!
    & \!\begin{tabular}{@{}c@{}}\vrule width 0pt depth 0pt height 8pt 
    {\small Arena}\\[-1pt]{\small Hard}\end{tabular}\!
    & \!\!\begin{tabular}{@{}c@{}}\vrule width 0pt depth 0pt height 8pt 
    {\small Alpaca}\end{tabular}\!\!
    \\
    \midrule
    Llama3 8B Instruct & 8.25 & 8.78 & 8.06 & 8.37 & 7.48 & 8.59 &\!\!\!& 
    8.41 & 8.74 & 8.07  & 8.59 & 8.06 & 8.63\\
    \midrule
    \textbf{ALMA: Our aligned models}~ & & & & & &  &\!\!&&&\\[1pt]
    ~~\textbf{11th round}\, \scalebox{0.93}{(= final)} & 8.22 & 8.77 & 7.86 & 8.33 & 7.67 & 8.48 &\!\!\!&
    8.34 & 8.65 & 7.89 & 8.42 & 8.26 & 8.46\\
    \begin{tabular}{@{}l@{}}\vrule width 0pt depth 0pt height 8pt 
    ~~1st round\, \scalebox{0.9}{(= initial SFT)}\end{tabular}~~~ & 7.94 & 8.55 & 7.60 & 8.18 & 7.18 & 8.20
    &\!\!\!& 7.78 & 8.27 & 7.36 & 7.89 & 7.54 & 7.87\\
    \midrule
    Llama3 8B Base (3-shot)~~~  & 4.00 & 4.44 & 3.58 & 4.29 & 3.37 & 4.36 &\!\!\!&
    3.95 & 4.33 & 3.67 & 4.30 & 3.71 & 3.74 \\
    \bottomrule
    \end{tabular}
    }\vspace{-1mm}
    \caption{\textbf{Main result on alignment benchmarks}. Our aligned models (middle rows) significantly improve on the base model (bottom row) and achieve a close performance to Llama3 Instruct (top row) across diverse alignment benchmarks---LIMA, MT-Bench, SR, Arena Hard, and Alpaca---, despite using less than 1\% of human annotations.
    For details on our progression from 1st to 11th round alignment, see Figure \ref{fig:result_curve}.
    }
    \label{tab:main_result}
\end{table}

\begin{figure}[h!]
    \vspace{3pt}
    \centering
    \includegraphics[width=0.45\linewidth]{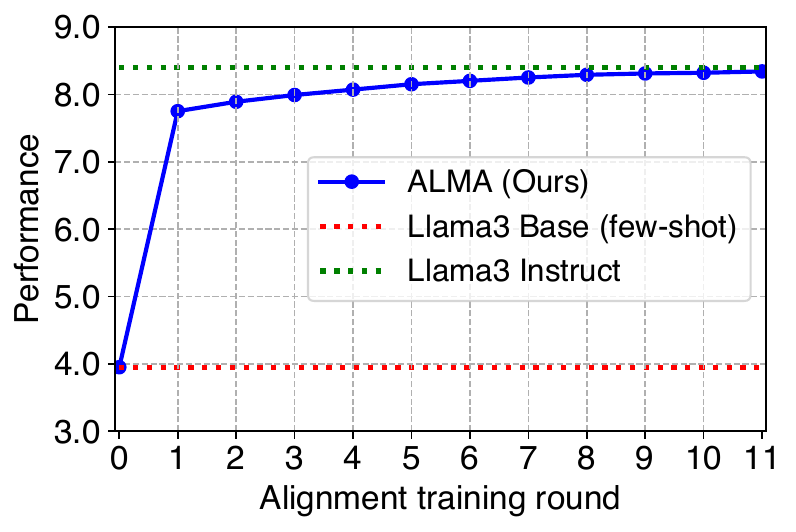}~~~~~~
    \vspace{-5pt}
    \caption{\textbf{Performance progression} of our alignment training, from 0th round (base model) to 1st round (initial SFT) to 11th round (our final model).
    The evaluation metric is the Armo score across alignment benchmarks: LIMA, MT-Bench, SR, Arena Hard, and Alpaca.
    We can see that starting from Llama3-Base (green line) and only using less than 10k labeled seed data, our model (blue line) steadily approaches the performance of Llama3-Instruct (red line).
    }
    \label{fig:result_curve}
\end{figure}
\begin{table}[t]
    \newcolumntype{G}{>{\columncolor[gray]{0.9}}r} 

    \centering
    \scalebox{0.85}{
    \small
    \begin{tabular}{lG}
    \toprule
    \begin{tabular}{@{}l@{}}\vrule width 0pt depth 0pt height 8pt 
    \textbf{Model}\end{tabular}
    & \begin{tabular}{@{}l@{}}\vrule width 0pt depth 0pt height 8pt 
    \textbf{Win Rate}\end{tabular}
    \\
    \midrule
    Llama3 8B Instruct & {22.9\%} \\
    \textbf{ALMA: Our final model} & {22.8\%} \\
    Baseline: Llama3 8B Base (3-shot) & 1.1\% \\
    \midrule
    \!\!\textit{(Models of similar sizes from the leaderboard)}~~~&\\[1pt]
    Mistral 7B v0.3 & 20.6\% \\
    GPT 3.5 Turbo 1106 & 19.3\% \\
    Llama2 Chat 70B & 14.7\% \\
    Gemma Instruct 7B & 10.4\% \\
    Alpaca 7B & 5.9\%\\
    \bottomrule
    \end{tabular}
    }\vspace{-1mm}
    \caption{ \textbf{AlpacaEval 2.0 results (length-controlled win rate over GPT-4 Turbo)}. Our aligned model (ALMA) achieves competitive performance to Llama3 Instruct in length-controlled evaluation metrics too. 
    }
    \label{tab:result_alpaca2}
\end{table}

Table \ref{tab:main_result} shows our main result.
Our aligned models (middle rows) significantly improve over the base model (bottom row) and achieve a close performance to Llama3 Instruct (top row) across diverse alignment benchmarks (LIMA, MT-Bench, SR, Arena Hard, and Alpaca), despite using less than 1\% of the human annotations. This is observed in both GPT-4o and Armo metrics.

Figure \ref{fig:result_curve} illustrates the performance progression from 0th round (base model) to 1st round (initial SFT) to 11th round (our final aligned model). 
The visualization shows how our model (blue line), starting from Llama3-Base (green line), consistently approaches toward Llama3-Instruct's performance (red line) over the 11 rounds.

These findings suggest that the base model can effectively bootstrap from minimal seed data ($<$1\% of the conventional amount) and generate high-quality synthetic data that replaces the extensive human-annotated data traditionally required for alignment.

\section{Analysis}
\label{sec:analysis}
In this section, we analyze the efficacy of our proposed techniques (\S \ref{sec:method-prompt}--\ref{sec:method-train}) in detail, including the judge development, prompt synthesis, response synthesis and training.

\subsection{Judge}
\begin{table}[t]
    \newcolumntype{G}{>{\columncolor[gray]{0.9}}c} 

    \centering
    \scalebox{0.85}{
    \small
    \begin{tabular}{lGcccc}
    \toprule
    \begin{tabular}{@{}l@{}}\vrule width 0pt depth 0pt height 8pt 
    \textbf{Judge}\end{tabular}
    & \begin{tabular}{@{}l@{}}\vrule width 0pt depth 0pt height 8pt 
    ~~~~Avg.~~~~\end{tabular}
    & \begin{tabular}{@{}l@{}}\vrule width 0pt depth 0pt height 8pt 
    ~~Chat~~\end{tabular}
    & \begin{tabular}{@{}l@{}}\vrule width 0pt depth 0pt height 8pt 
    Chat-Hard\end{tabular}
    & \begin{tabular}{@{}l@{}}\vrule width 0pt depth 0pt height 8pt 
    Safety\end{tabular}
    & \begin{tabular}{@{}l@{}}\vrule width 0pt depth 0pt height 8pt 
    Reasoning\end{tabular}
    \\
    \midrule
    \begin{tabular}{@{}l@{}}\vrule width 0pt depth 0pt height 8pt 
    \textbf{Our final method:}\\
    {Finetuned judge (real-value scoring) + self-distillation}
    \end{tabular} ~~~~~~& \textbf{0.863} & 0.922 & 0.537 & 0.870 & 0.924    \\
    \midrule
    Finetuned judge (real-value scoring) ~~~~ & 0.846 & 0.933 & 0.550 & 0.870 & 0.889 \\
    Finetuned judge (integer scoring) ~~~~ & 0.669 & 0.777 & 0.357 & 0.719 & 0.703    \\
    \midrule
    Baseline: 3-shot prompt Llama3-Base (integer scoring)  ~~~~& 0.492 & 0.601 & 0.372 & 0.627 & 0.447\\
    \bottomrule
    \end{tabular}
    }\vspace{-1mm}
    \caption{\small \textbf{Effect of our judge improvement techniques (\S \ref{sec:method-judge})}. Metric is judge accuracy on RewardBench. All of our proposed techniques---finetuning with high-quality seed data, real-value score prediction, and self-distillation---provide performance gains. Our final method that combines all performs the best (top row).
    }
    \label{tab:ablation_judge}
\end{table}

We introduced three main techniques to improve the judge: finetuning with high-quality seed data (\S \ref{sec:method-judge-data}), real-value scoring instead of integer scores (\S \ref{sec:method-judge-score}), and self-distillation (\S \ref{sec:method-judge-distill}).
Table \ref{tab:ablation_judge} shows the ablation study result. The evaluation metric is judge accuracy on RewardBench. The baseline judge uses few-shot prompting of the base LLM to produce integer scores from 0--10 (bottom row). Finetuning, real-value scoring, and self-distillation each provide substantial improvements in judge performance. Interestingly, self-distillation improved the performance on the reasoning category.
Our final method, which combines all these techniques, performs the best, achieving a score of 0.863 on RewardBench. 

\subsection{Prompt synthesis}
We analyze the efficacy of our prompt diversification technique, introduced in \S \ref{sec:method-prompt}.
Table \ref{tab:ablation_prompt} compares the alignment results between two approaches: using 50k random prompts (baseline) vs 50k diversified prompts (our final method using k-means clustering and subsampling). The diversified prompts consistently improved performance across all alignment benchmarks, with particularly notable gains on challenging ones that involve more reasoning-intensive and longer prompts, such as MT-Bench and Arena-Hard. This is likely because random sampling tends to over-represent common prompts, whereas our clustering approach ensures better coverage of the entire distribution, including complex and tail prompts.

\begin{table}[t]
    \newcolumntype{G}{>{\columncolor[gray]{0.9}}c} 

    \centering
    \scalebox{0.85}{
    \small
    \begin{tabular}{lGccccc}
    \toprule
    \begin{tabular}{@{}l@{}}\vrule width 0pt depth 0pt height 8pt 
    \textbf{Model}\end{tabular}
    & \begin{tabular}{@{}l@{}}\vrule width 0pt depth 0pt height 8pt 
    Avg.\end{tabular}
    & \begin{tabular}{@{}l@{}}\vrule width 0pt depth 0pt height 8pt 
    LIMA\end{tabular}
    & \begin{tabular}{@{}l@{}}\vrule width 0pt depth 0pt height 8pt 
    MT-Bench\end{tabular}
    & \begin{tabular}{@{}l@{}}\vrule width 0pt depth 0pt height 8pt 
    SR\end{tabular}
    & \begin{tabular}{@{}l@{}}\vrule width 0pt depth 0pt height 8pt 
    Arena-Hard\end{tabular}
    & \begin{tabular}{@{}l@{}}\vrule width 0pt depth 0pt height 8pt 
    Alpaca\end{tabular}
    \\
    \midrule
    \begin{tabular}{@{}l@{}}\vrule width 0pt depth 0pt height 8pt 
    {Trained with diversified prompts}\\
    \textbf{(our final method)}
    \end{tabular} ~~~~& \textbf{7.89} & \textbf{8.34} & \textbf{7.47} & \textbf{8.01} & \textbf{7.66} & \textbf{7.97}
    \\
    \midrule
    Trained with random prompts~~~~ & 7.79 & 8.30 & 7.23 & 7.97 & 7.53 & 7.92
    \\
    \bottomrule
    \end{tabular}
    }\vspace{-1mm}
    \caption{\small \textbf{Effect of our prompt diversification technique (\S \ref{sec:method-prompt})}. We compare the use of random prompts and diversified prompts during training and show the model performance on alignment benchmarks. The metric used is the Armo score of generated responses.
    }
    \label{tab:ablation_prompt}
\end{table}

\subsection{Response synthesis \& Training}
\label{sec:analysis-response}
Here, we analyze the effects of response sampling size and diversity techniques. 
We then discuss the hyperparameter choice in our training recipe, and how our recipe can be adapted to various preference optimization algorithms such as DPO.

\paragraph{Effect of response sampling size ($\bm{N}$).}
\begin{figure}[h!]
    \centering
    \includegraphics[width=0.45\linewidth]{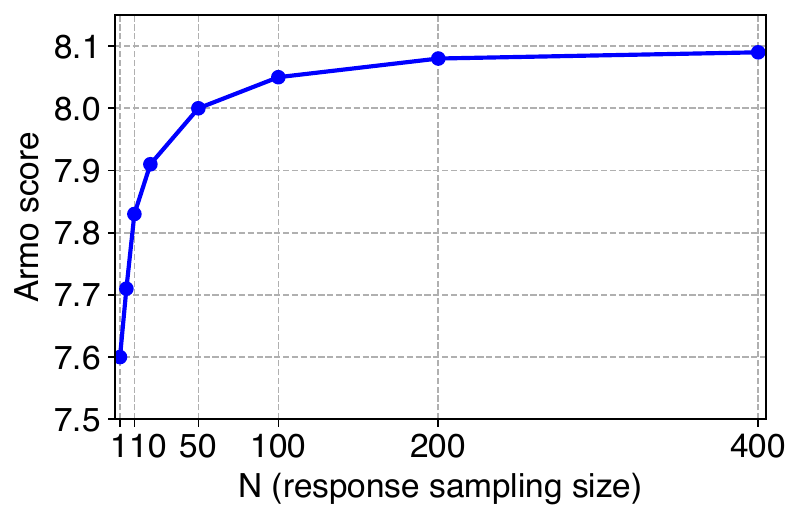}~~~~~~
    \vspace{-5pt}
    \caption{\small \textbf{Effect of response sampling size N on the quality of best-of-n response}. We evaluate the extrinsic reward score (Armo score) of the best response selected by our judge from N response samples, and analyze how performance scales with N. 
    While conventional alignment methods typically sample 10--20 responses, we find that the quality of the best-of-n response continues to improve beyond N=10 and even at N=200. Based on this result, we set N=200 for response synthesis in our alignment recipe (\S \ref{sec:method-response}).
    }
    \label{fig:best_of_n}
\end{figure}

Figure \ref{fig:best_of_n} visualizes the effect of response sampling size N on the quality of best-of-n response. Specifically, we evaluate the extrinsic reward score (Armo score) of the best response selected by our judge from N response samples, and plot how performance scales as we increase N from 1 to 10 and up to 400. 
While conventional rejection sampling methods typically sample 10--20 responses per prompt, we find that scaling N further is very effective:  the quality of the best-of-n response continues to improve beyond N=10 and even at N=200. Based on this result, we set N=200 for response synthesis in our alignment recipe (\S \ref{sec:method-response}).

\paragraph{Effect of diverse response sampling.}
\begin{table}[t]
    \newcolumntype{G}{>{\columncolor[gray]{0.9}}c} 
    \centering
    \scalebox{0.8}{
    \small
    \begin{tabular}{lGcccccccc}
    \toprule
    \multirow{2}{*}{\textbf{Method}} & \multicolumn{9}{c}{\textbf{MT-Bench}} \\
    \cmidrule{2-10}
    & Avg. & coding & \!\!extraction\!\! & \!\!humanities\!\! & math & \!\!reasoning\!\! & \!\!roleplay\!\! & stem & \!\!writing\!\!\\
    \midrule
    \begin{tabular}{@{}l@{}}\vrule width 0pt depth 0pt height 8pt 
    {Use the latest model and two SFT models}\\
    \textbf{(our final method)}\\\end{tabular}
    & \textbf{7.55} & \textbf{6.54} & \textbf{7.04} & \textbf{8.69} & \textbf{6.38} & \textbf{6.64} & \textbf{8.58} & \textbf{8.32} & \textbf{8.30}\\
    \midrule
    Only use the latest model & 7.53 & 6.52 & 7.02 & 8.65 & 6.34 & 6.63 & 8.56 & \textcolor{red}{8.26} & 8.29\\
    Only use the two SFT models & 7.51 & 6.43 & 7.00 & 8.64 & 6.31 & 6.61 & 8.51 & 8.30 & 8.27 \\
    Only use one SFT model & 7.48 & 6.41	&6.97 & 8.63 & 6.31 & 6.52 & 8.45 & 8.27 & 8.25 
    \\
    \bottomrule
    \end{tabular}
    }\vspace{-1mm}
    \caption{\small
    \textbf{Effect of diverse response sampling (\S \ref{sec:method-prompt})}. We experimented with using only the latest model ($M_{r-1}$), only the initial SFT models ($M_{1}$'s), or both types of models for response synthesis in the 3rd round ($r=3$) of best-of-n SFT process. For fair comparison, the total number of sampled responses per prompt ($N$) is fixed at 200 in all cases. We show the resulting model performance on MT-Bench, with a breakdown across all topic categories. We found that using only the latest model outperformed using only the initial SFT models on average, but it reinforced hallucination and underperformed on certain categories of prompts like STEM (shown in red in the table). We overcame this issue and achieved the best performance across all categories by using all models for response sampling, combining the strengths of each. 
    }
    \label{tab:ablation_multi}
\end{table}

In our iterative best-of-n SFT approach, we found that sampling responses from multiple model checkpoints outperforms using a single checkpoint for response synthesis. Our optimal configuration combines the latest aligned model ($M_{r-1}$) with two initial SFT models trained using different seeds ($M_{1}$'s), with a samping ratio of 1:1:1. This combination yields superior best-of-n SFT results compared to using any single model (see Table \ref{tab:ablation_multi}).

Our findings include:
\begin{itemize}
\setlength{\leftskip}{-26pt}
\item Using two SFT models generates more diverse responses compared to a single SFT model
\item The latest model and initial SFT models exhibit complementary strengths. While the latest model generally produces superior responses, the initial SFT models perform better on certain prompts. For example, when using only the latest model to sample responses for subsequent rounds of best-of-n SFT, we observed more hallucinated outputs and degraded performance on certain prompt categories, such as STEM (highlighted in red in Table \ref{tab:ablation_multi}). The initial SFT models maintain factual accuracy on these prompts due to their training on ground-truth seed data.
\item The initial SFT models generate more diverse responses than the latest model because they are relatively undertrained. Therefore, combining them with the latest model can maintain quality (via the latest model) while increasing response diversity (via the initial SFT models).
\end{itemize}
Thus, by leveraging both model types for response sampling, we achieved the best performance across all prompt categories.

\paragraph{Balancing response sampling size ($\bm{N}$), prompts ($\bm{K}$), and training rounds ($\bm{R}$) under fixed compute budget.}
\begin{table}[t]
    \newcolumntype{G}{>{\columncolor[gray]{0.9}}c} 

    \centering
    \scalebox{0.85}{
    \small
    \begin{tabular}{lG}
    \multicolumn{2}{c}{\textbf{Response sampling size (N) vs Number of prompts (K)}}\\[3pt]
    \toprule
    \textbf{Model}
    & \textbf{Eval score}   \\
    \midrule
    {N=200, K=50k} \textbf{(our final method)} ~~~~~& \textbf{7.89} \\
    \midrule
    N=400, K=25k~~~~ & 7.88\\
    N=100, K=100k~~~~ & 7.80\\
    \bottomrule
    \end{tabular}
    ~~~~~~
    \begin{tabular}{lG}
    \multicolumn{2}{c}{\textbf{Response sampling size (N) vs Number of rounds (R)}}\\[3pt]
    \toprule
    \textbf{Model}
    & \textbf{Eval score}   \\
    \midrule
    {N=200, R=10} \textbf{(our final method)} ~~~~~& \textbf{8.34} \\
    \midrule
    N=400, R=5~~~~ & 8.17\\
    N=133, R=15~~~~ & 8.13\\
    \bottomrule
    \end{tabular}

    }\vspace{-1mm}
    \caption{\small \textbf{Balancing response sampling size ($\bm{N}$), prompts ($\bm{K}$), and training rounds ($\bm{R}$) under fixed compute budget ($\bm{N \times K \times R}$)}. Under our current inference compute budget, we found optimal performance with the configuration of N=200 response samples, K=50k prompts, and R=10 training rounds.
    }
    \label{tab:ablation_config}
\end{table}

The total inference compute used in our alignment recipe can be expressed as $N \times K \times R$, where $N$ is the response sampling size per prompt, $K$ is the number of prompts per training round, and $R$ is the total number of training rounds. We investigated how the final model performance varies when adjusting $N$, $K$, and $R$ while keeping the total inference compute fixed.

Table \ref{tab:ablation_config} shows the results when we fix $R$ and vary $N$ and $K$ (left table) and the results  when we fix $K$ and vary $N$ and $R$ (right table), Under our current inference compute budget, we found that:
the effect of increasing \#prompts ($K$) starts to slow after 50k prompts; the effect of increasing \#rounds ($R$) starts to slow after 10 rounds. Hence, we selected our final values of $N$=200, $K$=50k, and $R$=10.
However, it is important to note that this analysis is specific to our current inference compute budget. With additional computational resources, model performance might continue to improve with larger values of $N$, $K$, and $R$.

\paragraph{Our alignment recipe can be used with various preference optimization algorithms (best-of-n SFT and DPO).}
\begin{table}[t]
    \newcolumntype{G}{>{\columncolor[gray]{0.9}}c} 
    \centering
    \scalebox{0.8}{
    \begin{tabular}{lGG}
    \toprule
    \multirow{2}{*}{\textbf{Model}} & \multicolumn{2}{c}{\textbf{Eval score}} \\
    \cmidrule{2-3}
    & ~~~~~~\textbf{DPO}~~~~~~ & \textbf{best-of-n SFT} \\
    \midrule
    6th round alignment & 8.16 & 8.20\\
    5th round alignment & 8.12 & 8.15\\
    4th round alignment & 8.07 & 8.07\\
    3rd round alignment & 8.03 & 7.99\\
    2nd round alignment & 7.99 & 7.89\\
    \midrule
    Initial SFT (=1st round)~~~~ & \multicolumn{2}{G}{7.78}\\
    \bottomrule
    \end{tabular}
    }\vspace{-1mm}
    \caption{\small
    \textbf{Various preference optimization objectives} can be used effectively with our alignment recipe, including best-of-n SFT and DPO.
    }
    \label{tab:ablation_obj}
\end{table}

So far, we have used best-of-n SFT in our experiments for its simplicity. Here we show that can work with other preference optimization methods, including the popular one, DPO \citep{rafailov2023dpo}.  
First, there are several design options for DPO when constructing training data (chosen response, rejected response) from response samples and judge scores. For example, we can (1) pick the best and worst responses; or (2) pick the best and random response; or (3) pick two random responses and order them. We found that (2) worked the best. So we adopt this DPO method in the rest of the experiment. 
Table \ref{tab:ablation_obj} shows the result when we use DPO in our alignment recipe in place of best-of-n SFT. We find that DPO also achieves continual improvement for 5+ rounds, like best-of-n SFT. 

Note that the focus of this paper is to present a general alignment recipe, which can be adapted with both best-of-n SFT and DPO (and potentially others too). Our focus is not on comparing best-of-n SFT vs DPO. 

\section{Conclusion}
This work introduced ALMA, a method that achieves effective language model alignment with minimal human annotation by leveraging the base model's inherent capabilities. Through careful data synthesis techniques---including diverse prompt generation, extensive response sampling, and judge enhancement via self-distillation---we show that a base model can bootstrap high-quality training data from just 9k labeled examples. Our approach achieves close performance to Llama3-Instruct while using less than 1\% of the human annotations used by conventional alignment methods. These findings can help democratize language model development by significantly reducing data annotation requirements.

\section{Limitations and future work}
In our evaluation, we mainly focused on established alignment benchmarks, such as MT-Bench, Alpaca, and Arena-Hard. While they cover diverse tasks and are representative of current alignment literature, we do not claim our model matches Llama3 Instruct across all possible tasks beyond typical alignment scenarios. In future, our method could be extended to incorporate new capabilities by including relevant examples in our seed SFT data, which is an interesting research direction. 
Additionally, while our current work focused on general alignment in terms of helpfulness and accuracy, future studies could explore safety alignment in greater depth.

In this study, we focused on the setting with 9k seed labeled examples (5k SFT and 4k judge samples). Future research could investigate the impact of both reducing and increasing the volume of seed data on model performance. Furthermore, future research could examine which types of seed human annotations are most valuable for bootstrapped data synthesis and which are indispensable (i.e., inherently difficult to synthesize). This would help better understand the complementary roles of synthetic and human-annotated data.

Lastly, our approach uses more inference compute than previous work during data synthesis, particularly for response synthesis. However, this tradeoff allows us to significantly reduce the amount of required human-annotated data.

\section*{Acknowledgments}
We thank Omer Levy, Gabriel Synnaeve, Scott Yih, Mary Williamson, Carleigh Wood, Lili Yu, Jacob Kahn, Kushal Tirumala, Arun Babu, and Michael Xie for their valuable feedback and support.

\bibliography{main}
\bibliographystyle{iclr2024_conference}

\appendix
\section{Judge}
\subsection{Annotation guideline for judge data}
\label{sec:app-judge-annotation}

\begin{lstlisting}[frame=single,breaklines=true,basicstyle=\small\ttfamily]
You are tasked with evaluating an AI assistant's performance on a user-defined task.

You are given a candidate response. Your task is to evaluate the quality of the response as an answer to the user input. Select the final answer from one of the following options:
0: Unusable - The answer does not provide any helpful or correct information.
1: Extremely Poor - Minimal relevance, major factual errors, or off-topic response.
2: Very Poor - Some relevance but highly inaccurate or lacking in substance.
3: Poor - Relevant but with significant gaps or major mistakes in understanding.
4: Below Average - Basic understanding but with notable inaccuracies or insufficient details.
5: Average - Provides a somewhat useful answer but lacks clarity, depth, or has minor inaccuracies.
6: Satisfactory - Generally clear and relevant but may miss some key points or have some minor issues.
7: Good - The answer is useful and mostly accurate with only minor areas for improvement.
8: Very Good - Clear, relevant, and mostly comprehensive, with only small gaps or omissions.
9: Excellent - Well-constructed, insightful, and highly accurate with only very minor room for improvement.
10: Outstanding - Perfect or near-perfect response; fully accurate, comprehensive, and insightful.

Please answer using one of these options verbatim.


[START USER INPUT]
<PROMPT>
[END USER INPUT]


[START ASSISTANT RESPONSE]
<RESPONSE>
[END ASSISTANT RESPONSE]


**What is your assessment of the response?**
Answer: <ANSWER>
\end{lstlisting}

\subsection{Prompt template for LLM-as-a-judge}
\label{sec:app-judge-prompt}

\begin{lstlisting}[frame=single,breaklines=true,basicstyle=\small\ttfamily]
Please act as an impartial judge and evaluate the quality of the response provided by an AI assistant to the user question displayed below. Your evaluation should consider factors such as the helpfulness, relevance, accuracy, depth, creativity, and level of detail of the response. Please just rate the response on a scale of 0 to 10 by strictly following this format: \"[[rating]]\", for example: \"Rating: [[3]]\".

# Prompt:
{prompt}

# Response:
{response}
\end{lstlisting}

\end{document}